\documentclass{article}

\usepackage{microtype}
\usepackage{xcolor}
\usepackage{graphicx}
\usepackage{subcaption}
\usepackage{booktabs} 
\usepackage{enumitem}
\usepackage{hyperref}

\usepackage[accepted]{icml2026}

\usepackage{amsmath}
\usepackage{amssymb}
\usepackage{mathtools}
\usepackage{amsthm}

\usepackage[capitalize,noabbrev]{cleveref}

\theoremstyle{plain}

\theoremstyle{definition}

\theoremstyle{remark}

\usepackage[disable,textsize=tiny]{todonotes}

\icmltitlerunning{Stop Thinking, Start Looking}

\begin{document}

\twocolumn[
  \icmltitle{Stop Thinking, Start Looking: Efficient Post-Training for Multimodal Document Question Answering via Reasoning-Free Alignment}




  \begin{icmlauthorlist}
\icmlauthor{Harikrishnan P M}{ind}
  \icmlauthor{Goutham Vignesh}{ind}
  \icmlauthor{Ganesh Parab}{ind}
  \icmlauthor{Saisubramaniam Gopalakrishnan}{ind}
  \icmlauthor{Vishal Vaddina}{can}
  \icmlauthor{Varun V}{ind}
  \icmlauthor{Rohit Agrawal}{ind}
  \end{icmlauthorlist}

\icmlaffiliation{ind}{Phi Labs, Quantiphi, India}
  \icmlaffiliation{can}{Phi Labs, Quantiphi, Canada}

  \icmlcorrespondingauthor{Harikrishnan P M}{harikrishnan.m@quantiphi.com}

  \icmlkeywords{Document Visual Grounding, Reinforcement Learning, Multimodal QA, GRPO, Reasoning}

  \vskip 0.3in
]



\printAffiliationsAndNotice{}  

\begin{abstract}
Efficient multimodal document question answering with explicit visual grounding, locating the precise document region that supports each answer remains an open challenge. Current approaches bifurcate into Supervised Fine-Tuning (SFT), which requires large annotated datasets and reaches optimization plateaus, and reasoning-centric Reinforcement Learning (RL), which depends on verbose intermediate traces that inflate inference token cost without clear benefit. We introduce \textbf{Perception-RFT}, a training framework that applies Group Relative Policy Optimization (GRPO) to multimodal document QA, bypassing intermediate reasoning tokens to directly align visual features with structured grounding outputs. To rigorously evaluate the necessity of reasoning, we construct a reasoning variant under identical reward settings. We find that reasoning-enabled models suppress their reasoning traces during training, converging to direct perception-based policies at the 4B parameter scale, reducing per-query inference token length by more than 60\%, while reasoning-enabled RL underperforms perception-only training. Through a fine-grained analysis of Qwen3-VL-4B optimization dynamics, we confirm that SFT saturation and cold-start RL instability established in text-domain post-training extend to multimodal, and identify a previously uncharacterized \textbf{Grounding Divergence}: a selective trade-off between semantic robustness and geometric precision on two out of distribution (OOD) benchmarks (4,828 samples) under joint RL optimization. We further show that an early SFT$\rightarrow$RL transition achieves comparable precision with 65\% less training data.
\end{abstract}

\section{Introduction}
\label{sec:intro}

The multimodal question answering over documents requires more than just generating the correct text, it requires localizing the precise visual region that supports each answer \cite{11235566,pantazopoulos2025towards}. This capability, which we term \textbf{Document Visual Grounding (DVG)}, requires models to jointly perform semantic extraction and precise geometric localization from a single document image. Efficient solutions to DVG are practically critical: in high-stakes settings such as legal auditing and financial analysis, stakeholders need answers grounded in their source regions, not just text output \cite{sun2024review}. However, despite strong textual understanding, MLLMs remain fragile at fine-grained spatial grounding under distribution shifts \cite{fu2024ocrbench,li2025towards}, and existing training strategies impose significant compute overhead without fully resolving this fragility.

To address this challenge, existing approaches have largely bifurcated into two paradigms. Supervised Fine-Tuning (SFT), exemplified by large-scale benchmarks such as DOGR \cite{zhou2025dogr}, relies on scaling annotated datasets to enforce generalization. In contrast, reasoning-centric Reinforcement Learning (RL) approaches \cite{guo2025deepseek}, such as DocThinker \cite{yu2025docthinker}, posit that accurate grounding requires explicit intermediate reasoning traces to bridge semantic understanding and spatial localization. However, the optimization dynamics underlying these paradigms is not yet sufficiently understood. \textbf{Does scaling SFT truly improve robustness, or does it introduce hidden trade-offs? Are verbose reasoning traces necessary for grounding, or do they impose unnecessary computational overhead for inherently perceptual tasks?}

We extend recent findings that challenge the view that ``SFT memorizes while RL generalizes'' \cite{chu2025sft,jin2025rl} to the multimodal QA domain. We hypothesize that DVG is a perception-dominant task, where direct alignment between visual features and spatial outputs is more effective than reasoning-mediated generation a view supported by recent studies showing that explicit reasoning introduces latency and instability for visual primitives \cite{yu2025perception,li2025think}.

To investigate this hypothesis, we introduce \textbf{Perception-RFT}, a training framework that applies Group Relative Policy Optimization (GRPO) \cite{shao2024deepseekmath} to the DVG domain, optimizing semantic and geometric correctness without requiring intermediate reasoning tokens. Crucially, rather than assuming that reasoning is unnecessary, we explicitly enable reasoning within the same RL framework to study its role during optimization. This allows us to perform a controlled comparison between reasoning-enabled and reasoning-free training under identical reward formulations.

Our analysis reveals a consistent behavioral pattern: when reasoning is enabled, models initially generate detailed reasoning traces, but progressively compress and ultimately eliminate them as training proceeds. At convergence, the learned policy produces minimal or no reasoning tokens while maintaining or improving DVG performance. Furthermore, reasoning-enabled RL does not improve performance and can underperform direct perception-based training, indicating that optimal grounding policies rely on direct perception rather than explicit reasoning.

Beyond this central finding, we provide a fine-grained analysis of SFT and RL dynamics for DVG. We confirm that, the two dynamics established in text-domain post-training \cite{guo2025deepseek} extend to multimodal grounding: supervised optimization saturates while RL continues to improve task-aligned behavior, and SFT initialization is critical for stable geometric learning under RL. We further identify a previously uncharacterized phenomenon, the \textbf{Grounding Divergence}: a selective trade-off in which localization improves while semantic robustness degrades on out-of-distribution documents under joint RL optimization. Finally, we show that an early transition from SFT to RL achieves comparable precision with significantly less supervised data.

\section{Related Work}
\label{sec:related}

\subsection{From OCR Pipelines to Native Perception}
DVG has relied on pipelined systems \cite{huang2022layoutlmv3,cui2025paddleocr} that fuse OCR tokens with spatial embeddings, a paradigm modern LLMs perpetuate via external OCR inputs \cite{openai20234v,team2024gemini,vishal2025drishtikon}. Empirical studies reveal a critical modality gap: without OCR, state-of-the-art models achieve near-zero IoU, and even Oracle OCR fails to resolve it. Models hallucinate boxes from text order rather than visual layout \cite{li2025towards,fu2024ocrbench}. Multi-stage workarounds such as DLaVA \cite{mohammadshirazi2024dlava} boost accuracy but introduce significant pipeline complexity. Our work eliminates these external dependencies by aligning the MLLM's internal features directly to coordinate regression.

\subsection{Visual Grounding and Benchmarks}
To enable native capability, recent works have integrated bounding box coordinates into the MLLM vocabulary. Generalist models like Shikra \cite{chen2023shikra}, Kosmos \cite{lv2023kosmos} and Ferret \cite{you2023ferret} demonstrated that MLLMs could output coordinates for natural images. In the document domain, DOGR \cite{zhou2025dogr} established a comprehensive benchmark, proposing a data engine to generate millions of supervised samples. However, these paradigms predominantly rely on SFT. We hypothesize that SFT optimization encounters a saturation point, where further training yields diminishing returns on geometric precision while potentially degrading semantic flexibility on out of distribution (OOD) documents.

\subsection{The Role of Reasoning in Vision Optimization}
Recent work adopts reasoning-intensive RL strategies for visual tasks \cite{liu2025visual,huang2025vision,shen2025vlm}. Reasoning-Centric models such as DocThinker \cite{yu2025docthinker} and Reason-RFT \cite{tan2025reason} demonstrate that explicit CoT reasoning traces benefit tasks with strong reasoning structure such as grounding via semantic decomposition and abstract visual problems like counting and geometric transformation. However, Perception-Centric studies \cite{yu2025perception,li2025think} show that such traces introduce latency and instability for visual primitives. Crucially, whether reasoning is retained or eliminated under RL in perception-dominant tasks and whether it helps or hurts remains underexplored. We investigate this directly by enabling reasoning within the same RL framework and analyzing its role during training.

\section{Methodology}
\label{sec:method}
\begin{figure*}[t]
\centering
\includegraphics[width=1.0\textwidth]{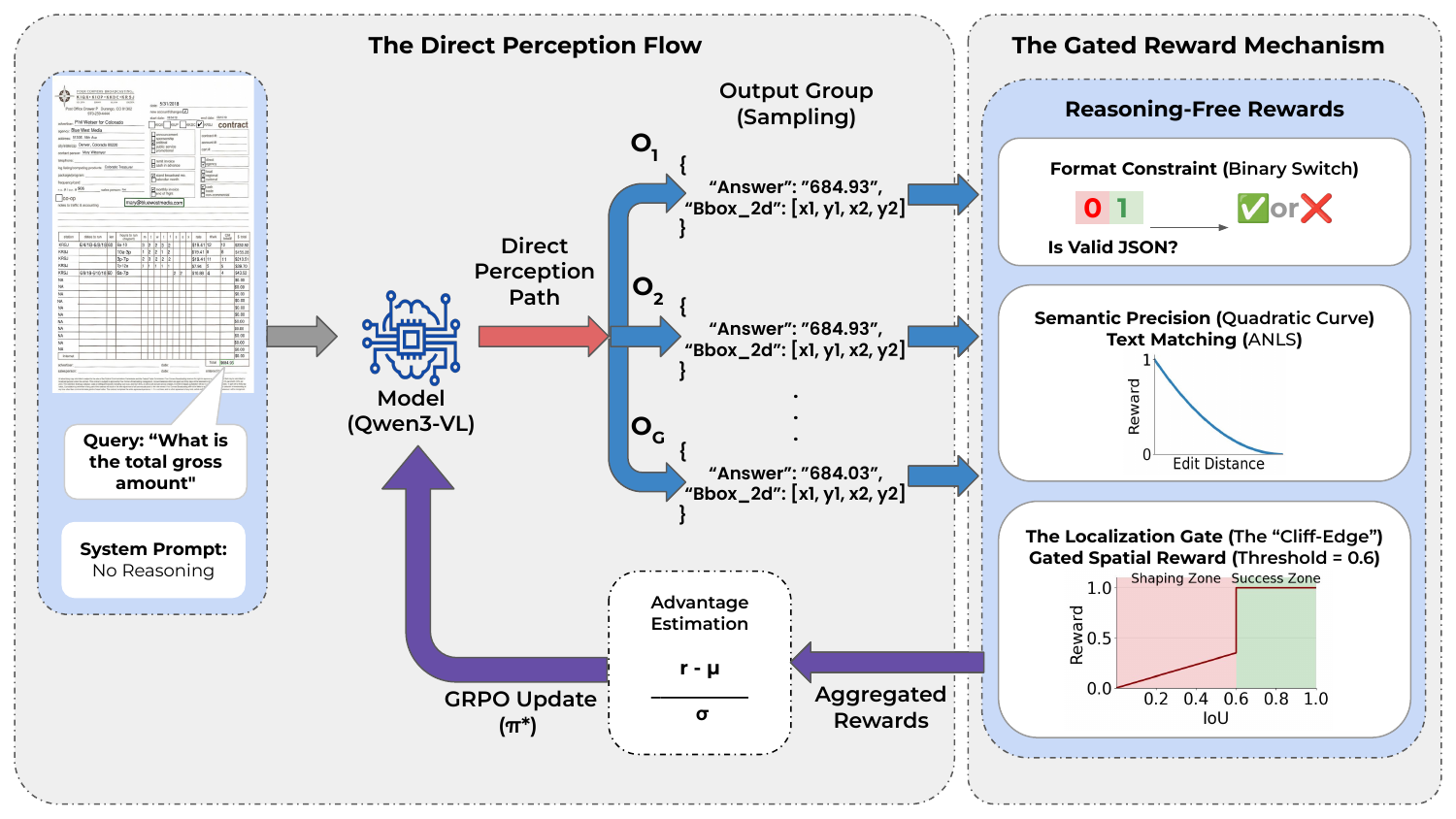}
\caption{\textbf{The Perception-RFT Framework.} Left: System prompting enforces the \textbf{Direct Perception} constraint. Right: The \textbf{Gated Dense Reward} mechanism stabilizes GRPO by defining separate shaping and success regions. \textbf{(Best viewed zoomed in.)}}
\label{fig:architecture}
\end{figure*}
\subsection{Task Formulation: Direct Perception}
We formulate the task of Document Visual Grounding (DVG) as a generative Visual QA task with explicit grounding. Given a single document image $d \in \mathbb{R}^{H \times W \times 3}$ and a text query $q$, the policy $\pi_\theta$ must generate a structured JSON output $Y$ containing the textual answer $A$ and its corresponding bounding box $B$:
\begin{equation}
  Y = \{ \texttt{"answer"}: A, \texttt{"bbox\_2d"}: [x_1, y_1, x_2, y_2] \}
\end{equation}
where coordinates are normalized to the range $[0, 1000]$ following the native Qwen3-VL architecture.

\noindent\textbf{The ``Direct Perception'' Constraint:} A critical deviation of our approach is the explicit suppression of reasoning. Unlike recent paradigms that rely on verbose intermediate traces (e.g. \texttt{<think>} tokens) to guide generation, we enforce a \textbf{Direct Perception} constraint via system prompting: \textit{``Do NOT include explanations, reasoning, or intermediate steps; Output ONLY the final valid JSON result.''} This forces the optimization process to bypass verbose thought chains and align visual features directly to coordinate regression outputs.

\subsection{Optimization Framework: Perception-RFT}
We optimize the policy $\pi_\theta$ using Group Relative Policy Optimization (GRPO) \cite{shao2024deepseekmath,guo2025deepseek}. As illustrated in Figure \ref{fig:architecture}, while GRPO is conventionally utilized to elicit latent reasoning, we apply it to enforce Direct Perception Alignment in the DVG domain, a configuration we refer to as \textbf{Perception-RFT}. By explicitly bypassing the reasoning module, we force the policy to map visual features directly to coordinate output.

For each query $q$, we sample a group of outputs $G = \{o_1, \dots, o_G\}$ and compute the advantage $A_i$ based on the group's normalized reward distribution. The objective maximizes the advantage of valid outputs while constraining policy drift:
\begin{equation}
  \mathcal{J}(\theta) = \mathbb{E}_{q \sim D} \left[ \frac{1}{G} \sum_{i=1}^{G} \left( \frac{\pi_\theta(o_i|q)}{\pi_{old}(o_i|q)} A_i - \beta D_{\mathrm{KL}}(\pi_\theta \| \pi_{ref}) \right) \right]
\end{equation}
By bypassing ``intermediate reasoning representations'', we shift the optimization landscape from maximizing  logical consistency  to maximizing \textbf{joint semantic and geometric precision}. This constraint is specific to the perception-only variant; a reasoning-enabled variant is introduced separately for controlled comparison.

\subsection{Gated Dense Perception Reward}
To enable precise visual grounding without the guidance of intermediate reasoning steps, we introduce a Gated Dense Reward mechanism. We hypothesize that standard dense rewards (e.g., IoU-based objectives employed in \cite{liu2025visual, shen2025vlm}) may inadvertently encourage models to predict ``safe'', oversized bounding boxes to maximize overlap coverage during early training stages. To proactively counteract this potential degeneracy, our function imposes a strict success gate that bifurcates the optimization surface.

The total reward $R$ is an equally weighted sum of three distinct objectives:
\begin{equation}
  R = R_{\text{format}} + R_{\text{sem}} + R_{\text{loc}}
\end{equation}

\noindent\textbf{1. Format Consistency ($R_{\text{format}}$):}
To ensure strict schema adherence for downstream parsing, we apply a binary reward. $R_{\text{format}} = 1$ if and only if the output is valid JSON containing strictly formatted \texttt{"answer"} and \texttt{"bbox\_2d"} keys. If parsing fails, this component is 0.

\noindent\textbf{2. Semantic Precision ($R_{\text{sem}}$):}
We utilize the Average Normalized Levenshtein Similarity (ANLS) \cite{anls} to evaluate textual accuracy while remaining robust to minor OCR character variations. To penalize near-misses more aggressively than standard F1, we apply a quadratic scaling:
\begin{equation}
  R_{\text{sem}} = 
  \begin{cases} 
  1.0 & \text{if } \text{ANLS} = 1.0 \text{ (Exact Match)} \\
  \text{ANLS}(\text{pred}, \text{gt})^2 & \text{otherwise}
  \end{cases}
\end{equation}

\noindent\textbf{3. Gated Localization Reward ($R_{\text{loc}}$):}
This component prevents the model from converging to suboptimal overlaps. We define a Gated IoU function:
\begin{equation}
  R_{\text{loc}} = 
  \begin{cases} 
  1.0 & \text{if } \text{IoU} \ge \tau \text{ (Success State)} \\
  \lambda \cdot \text{IoU} & \text{if } 0 < \text{IoU} < \tau \text{ (Shaping State)} \\
  0.0 & \text{otherwise}
  \end{cases}
\end{equation}
\textbf{Mechanism:} 
\begin{itemize}
\item \textbf{The Shaping State:} When alignment is poor ($\text{IoU} < \tau$), the reward is significantly scaled down. This provides just enough gradient signal to guide the model towards the target, but prevents it from converging on suboptimal partial overlaps. The value $\lambda=0.3$ is set empirically.
\item \textbf{The Success State:} We set the success threshold $\tau=0.6$ empirically, positioning it slightly above the standard evaluation metric ($\text{IoU} \ge 0.5$). This encourages the policy to learn a geometric safety margin, ensuring that predictions remain robust even under minor perturbations. Crossing this threshold triggers a discrete jump to maximum reward (1.0), creating a ``cliff-edge'' optimization surface.
\end{itemize}
We emphasize that semantic and geometric objectives are jointly optimized under a unified reward without explicit bias toward either component.
\subsection{Reasoning-Enabled Variant}
\label{sec:reasoning_variant}

To isolate the role of explicit reasoning, we construct a variant where the model may generate intermediate \texttt{<think>} traces before the structured JSON output. The reward function and optimization objective remain identical to Perception-RFT, with no supervision on the reasoning traces, enabling a controlled comparison under identical conditions.

\subsection{Dataset Construction}
To fuel this optimization, we prioritized high-fidelity supervision over scale. We curated a comprehensive DVG corpus by manually aggregating and adapting established key information extraction (KIE) benchmarks specifically, \textbf{DocILE} \cite{vsimsa2023docile} and \textbf{FormNLU} \cite{ding2023form}. We explicitly selected these curated sources over recent automated benchmarks such as BoundingDocs \cite{giovannini2025boundingdocs} to avoid the repeated token ambiguity inherent in OCR mapped datasets. Since our source datasets provide static Key-Value pairs, we implemented a task adaptation protocol to transform them into dynamic document visual QA samples. To ensure robustness, we employed a three-stage transformation strategy. First, we applied Semantic Augmentation \cite{tito2024privacy}, utilizing LLMs to generate diverse natural language aliases for canonical keys (e.g., mapping ``date\_issue'' to ``Invoice Date'') to enforce conceptual understanding over simple string matching. Second, we enforced Phrasing Generalization by using disjoint prompt templates for training and evaluation. Finally, we mitigated spatial overfitting through Dynamic Sub-Sampling, selecting a random subset (max 20\%) of available keys per epoch to ensure the model treats every iteration as a fresh visual search task rather than a layout memorization exercise.

\section{Experimental Setup}
\label{sec:setup}

\subsection{Datasets}
Our experiments utilize a carefully stratified dataset to rigorously test generalization across layout regimes.

\begin{itemize}
  \item \textbf{Training Set (General Financial Documents):} We aggregated DocILE and FormNLU to create a comprehensive training corpus of \textbf{23,696} unique QA samples. Far from being limited to simple invoices, this set covers a broad taxonomy of financial documents, including Tax Invoices, Purchase Orders, Receipts, Utility Bills, Credit/Debit Notes, and Tax Form 604s. This ensures the model learns a generalized representation of financial business document semantics and tabular structures.

  \item \textbf{In-Distribution (ID) Evaluation Set:} A strictly held-out subset of \textbf{6,194} samples from the same financial business domain as the training set, used to measure in-distribution precision.

  \item \textbf{OOD Evaluation Sets:} To assess generalization under distribution shift, we evaluate on two independent benchmarks. (1)~\textbf{DOGR-Bench} \cite{zhou2025dogr} (800 samples) comprises Crello artistic posters, ChartVQA bar graphs, and dense scientific PDFs visually and structurally distant from our financial training documents. (2)~\textbf{MMDocBench} \cite{mmdocbench} (4,028 samples; Key Information Extraction and Document QA subtasks) spans financial reports, receipts, scientific papers, and infographics. Together these benchmarks provide 4,828 OOD samples across diverse visual topologies.
\end{itemize}

\subsection{Models and Baselines}
We utilize \textbf{Qwen3-VL-4B} \cite{bai2025qwen3vltechnicalreport} as our foundation due to its strong native resolution handling. To demonstrate the accessibility of our approach, all experiments were conducted on a \textbf{single NVIDIA A100 (80GB)} using \textbf{Unsloth} \cite{unsloth2024}. We employed LoRA \cite{lora} ($r=16, \alpha=16$) for memory-efficient adaptation across all stages.

We evaluate the following configurations:

\begin{enumerate}
  \item \textbf{SFT (Supervised Fine-Tuning):} Qwen3-VL-4B fine-tuned for \textbf{3 full epochs} (1113 steps) on the 23k training set using the AdamW 8-bit optimizer. This represents the converged limit of standard supervised learning.

  \item \textbf{RFTs (SFT $\rightarrow$ RL):} Perception-RFT applied via continued training on the SFT model using a limited budget of \textbf{6k samples} ($<10\%$ of the SFT compute), with GRPO group size $G=8$.

  \item \textbf{RFTb (Cold-Start RL):} Perception-RFT applied directly to the base model, without any supervised initialization.

  \item \textbf{Reasoning-RFTb:} A reasoning-enabled variant of RFTb, where the model is allowed to generate intermediate reasoning traces (e.g., \texttt{<think>} tokens) before producing the final structured output. The reward function and optimization setup remain identical to RFTb.

\item \textbf{Gemini 3.0 Flash:} Evaluated zero-shot on the same image-question input as our models, serving as a frontier generalist reference point.
\end{enumerate}

We exclude DocThinker \cite{yu2025docthinker} from direct comparison as no code, model weights, or compatible evaluation artifacts are publicly available beyond the paper itself.

\subsection{Evaluation Metrics}
We report three strict metrics without text normalization:

\begin{itemize}
  \item \textbf{$\mathbf{F1}_{\textbf{EM}}$:} Measures the Exact Match token-level F1 of the answer string.

  \item \textbf{$\mathbf{F1}_{\textbf{loc}}$:} Defines visual grounding success strictly as $\text{IoU} \ge 0.5$.

  \item \textbf{$\mathbf{F1}_{\textbf{all}}$ (Strict Joint Success):} A compound metric requiring \textbf{both} perfect text extraction ($ \text{F1}_{\text{EM}} = 1.0$) and successful localization ($\text{IoU} \ge 0.5$), simulating enterprise auditing scenarios where partially correct answers are unacceptable.
\end{itemize}
All reported metrics are averaged over 3 independent runs evaluated on held-out sets of 6,194 (ID) and 4,828 OOD samples (DOGR-Bench: 800; MMDocBench: 4,028).
\section{Results and Analysis}
\label{sec:results}

\subsection{Conceptual Framework}

Our analysis centers on three dynamics: (1)~SFT optimization saturates beyond a point where cross-entropy training fails to improve grounding precision; (2)~SFT warm-start is required for stable geometric learning under RL, confirming cold-start instability established in prior work \citep{guo2025deepseek}; and (3)~the \textbf{Grounding Divergence}, a selective trade-off under distribution shift where localization gains are not matched by semantic robustness.
\subsection{Main Results: Generalization and Alignment}
\begin{table*}[t]
\centering
\caption{\textbf{Main results on ID and OOD evaluation sets.} OOD results span two independent benchmarks: DOGR-Bench (800 samples) and MMDocBench (4,028 samples). RFTs achieves the strongest ID joint grounding. Gemini 3.0 Flash is evaluated zero-shot under identical input conditions as our models. DOGR SFT is evaluated on DOGR-Bench only.}
\label{tab:main_results}
\resizebox{0.99\textwidth}{!}{%
\begin{tabular}{ll ccc ccc ccc}
\toprule
& & \multicolumn{3}{c}{\textbf{ID (Finance)}} & \multicolumn{6}{c}{\textbf{OOD}} \\
\cmidrule(lr){3-5}\cmidrule(lr){6-11}
\textbf{Model} & \textbf{Setup} & $\mathbf{F1}_{\textbf{EM}}$ & $\mathbf{F1}_{\textbf{loc}}$ & $\mathbf{F1}_{\textbf{all}}$ & \multicolumn{3}{c}{\textbf{DOGR-Bench}} & \multicolumn{3}{c}{\textbf{MMDocBench}} \\
\cmidrule(lr){6-8}\cmidrule(lr){9-11}
& & & & & $\mathbf{F1}_{\textbf{EM}}$ & $\mathbf{F1}_{\textbf{loc}}$ & $\mathbf{F1}_{\textbf{all}}$ & $\mathbf{F1}_{\textbf{EM}}$ & $\mathbf{F1}_{\textbf{loc}}$ & $\mathbf{F1}_{\textbf{all}}$ \\
\midrule
\multicolumn{11}{l}{\textit{Internal Baselines}} \\
Qwen3-VL-4B & Zero-Shot & 0.558 & 0.324 & 0.262 & 0.743 & 0.416 & 0.359 & 0.636 & 0.496 & 0.389 \\
Qwen3-VL-4B & SFT & 0.756 & 0.769 & 0.668 & 0.722 & 0.736 & 0.666 & 0.616 & 0.685 & 0.555 \\
\textbf{Qwen3-VL-4B} & \textbf{RFTs} & \textbf{0.773} & \textbf{0.821} & \textbf{0.718} & 0.722 & 0.759 & 0.685 & 0.620 & 0.712 & 0.569 \\
Qwen3-VL-4B & RFTb & 0.601 & 0.496 & 0.411 & 0.732 & 0.667 & 0.600 & 0.702 & 0.673 & 0.552 \\
\midrule
\multicolumn{11}{l}{\textit{External References}} \\
Gemini 3.0 Flash & Zero-Shot & 0.658 & 0.657 & 0.581 & 0.801 & 0.765 & 0.715 & \textbf{0.766} & \textbf{0.869} & \textbf{0.701} \\
DOGR \citep{zhou2025dogr} & SFT & {--} & {--} & {--} & \textbf{0.832} & {--} & \textbf{0.730} & {--} & {--} & {--} \\
\bottomrule
\end{tabular}
}
\end{table*}
We evaluate reinforcement learning for DVG across three settings: ID (Finance), and two OOD benchmarks, DOGR-Bench and MMDocBench. Table~\ref{tab:main_results} compares the base model, supervised fine-tuning (SFT), and reinforcement fine-tuning under two regimes: \textbf{RFTs} and \textbf{RFTb}. External models are included only as reference points.

\paragraph{Beyond supervised saturation.}
On ID data, \textbf{SFT} establishes a strong baseline ($F1_{\text{all}}=0.668$), but \textbf{RFTs} improves performance further to $0.718$. Gains are observed in both extraction ($0.756 \rightarrow 0.773$) and localization ($0.769 \rightarrow 0.821$), indicating that reinforcement learning refines both semantic and geometric alignment beyond what likelihood-based training achieves. This confirms that supervised learning saturates while policy optimization continues to improve task-aligned behavior, consistent with known limitations of likelihood-based training.

\paragraph{Importance of supervised initialization.}
Comparing \textbf{RFTs} and \textbf{RFTb} highlights the role of supervised initialization. While \textbf{RFTb} improves over the base model, it remains substantially below \textbf{RFTs} on both ID and OOD. This indicates that SFT provides a stable initialization that enables more effective downstream optimization under RL, confirming the cold-start instability finding of DeepSeek-R1 \citep{guo2025deepseek} in the multimodal document QA domain.

\paragraph{Generalization under distribution shift.}
Across both OOD benchmarks, \textbf{RFTs} achieves the highest joint grounding ($F1_{\text{all}}$: DOGR $0.685$, MMDocBench $0.569$) and the strongest localization gains over the base model (DOGR: $0.416 \rightarrow 0.759$; MMDocBench: $0.496 \rightarrow 0.712$). Semantic extraction, however, tells a more nuanced story. On DOGR-Bench, all trained variants fall below the zero-shot base ($0.722$ vs.\ $0.743$) a clear Grounding Divergence. On MMDocBench, \textbf{RFTb} ($0.702$) exceeds the base ($0.636$), while SFT ($0.616$) and RFTs ($0.620$) fall below it. This suggests that cold-start RL, by avoiding SFT over-specialization, better preserves semantic flexibility and that the severity of the Grounding Divergence varies with visual domain distance. This pattern is structurally distinct from catastrophic forgetting, where all capabilities degrade uniformly: here, geometric precision transfers robustly to unseen domains while semantic extraction does not, indicating that RL optimization induces asymmetric generalization across task dimensions. In both cases, the net effect on joint grounding is positive, confirming that localization gains outweigh semantic regression as an operationally critical metric.

\paragraph{Scale versus alignment.}
Comparing against Gemini 3.0 Flash under identical zero-shot conditions reveals a clear specialist-vs-generalist trade-off. On ID, our task-aligned RFTs significantly outperforms Gemini 3.0 Flash ($F1_{\text{all}}$: $0.718$ vs $0.581$), demonstrating that RL-based alignment to the target task yields superior in-domain precision over a much larger frontier model. On DOGR-Bench, the gap narrows ($0.685$ vs $0.715$), and on the broader MMDocBench, Gemini's generalist training gives it an advantage ($0.569$ vs $0.701$)  expected given our model was trained exclusively on financial documents while Gemini was exposed to diverse visual domains at scale. This confirms that task alignment is more critical than scale within the training domain, while generalist models retain a natural advantage on broader cross-domain evaluation. Crucially, the DOGR baseline comparison requires careful interpretation: the DOGR model was trained on a large-scale corpus explicitly sourced from the same visual domains as DOGR-Bench (artistic posters, charts, scientific PDFs), making the benchmark effectively in-distribution for that model. Our model, trained exclusively on financial documents, treats DOGR-Bench as a genuine cross-domain test. Our 0.685 $F1_{\text{all}}$ therefore reflects true out-of-distribution generalization rather than held-out performance within the training distribution.
\subsection{Optimization Dynamics: Stability and Efficiency}

\begin{figure*}[t]
  \centering
  \begin{subfigure}[b]{0.32\textwidth}
    \centering
    \includegraphics[width=\textwidth]{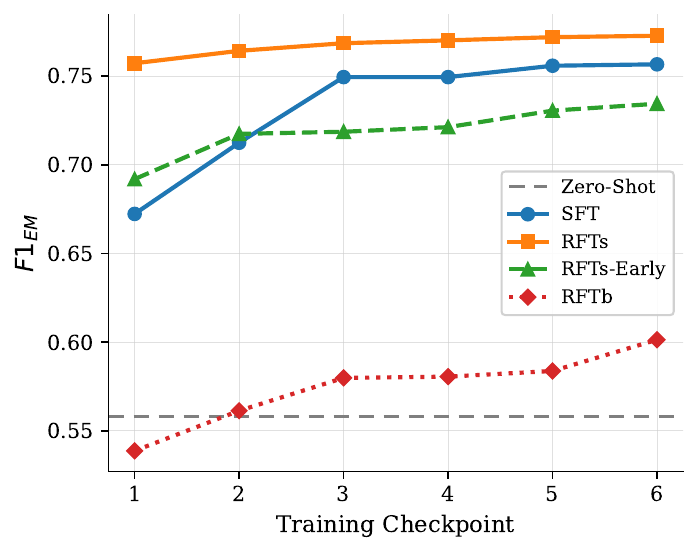}
    \caption{}
    \label{fig:ID_em}
  \end{subfigure}
  \hfill
  \begin{subfigure}[b]{0.32\textwidth}
    \centering
    \includegraphics[width=\textwidth]{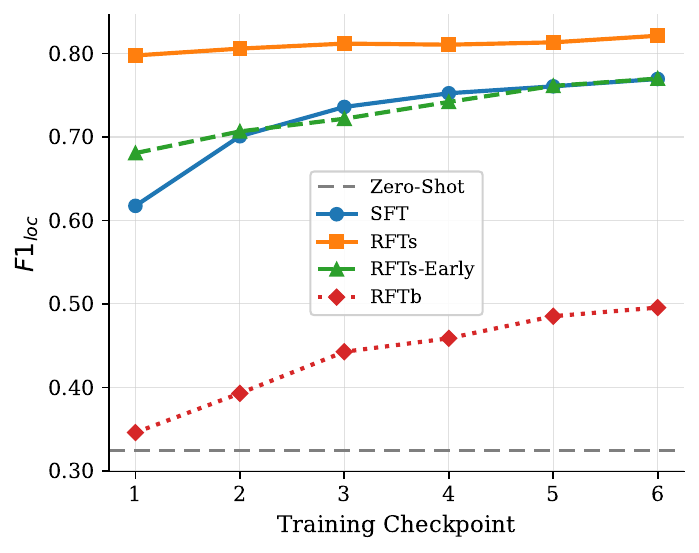}
    \caption{}
    \label{fig:ID_loc}
  \end{subfigure}
  \hfill
  \begin{subfigure}[b]{0.32\textwidth}
    \centering
    \includegraphics[width=\textwidth]{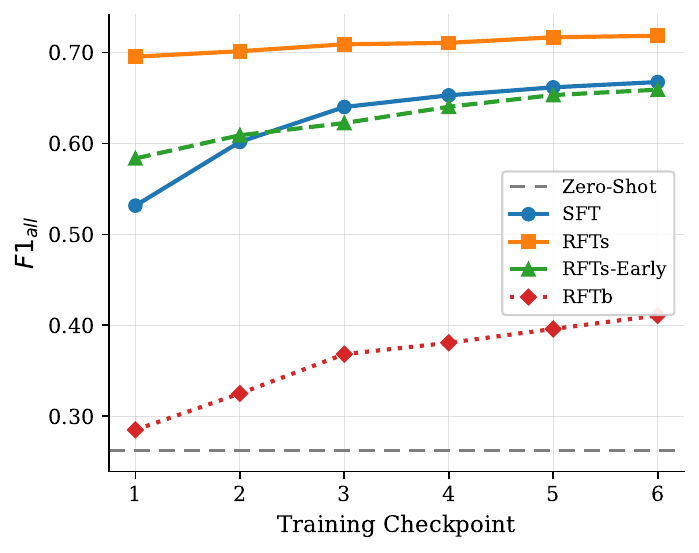}
    \caption{}
    \label{fig:ID_all}
  \end{subfigure}
  
  \caption{\textbf{ID Optimization Dynamics.} \textbf{SFT} (Blue) plateaus early across all metrics. \textbf{RFTs} (Orange) breaks the ceiling. Notably, in (b), \textbf{RFTs-Early} (Green) eventually \textbf{matches or slightly exceeds the localization performance of Full SFT}, proving that early-exit policy optimization is more data-efficient than prolonged supervised learning. \textbf{(Best viewed zoomed in.)}}
  \label{fig:id_dynamics}
\end{figure*}

\begin{figure*}[t]
  \centering
  \begin{subfigure}[b]{0.32\textwidth}
    \centering
    \includegraphics[width=\textwidth]{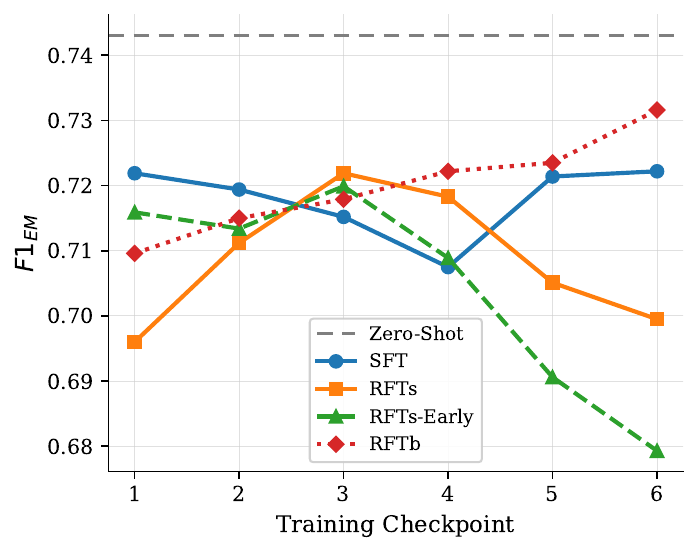}
    \caption{}
    \label{fig:ood_em}
  \end{subfigure}
  \hfill
  \begin{subfigure}[b]{0.32\textwidth}
    \centering
    \includegraphics[width=\textwidth]{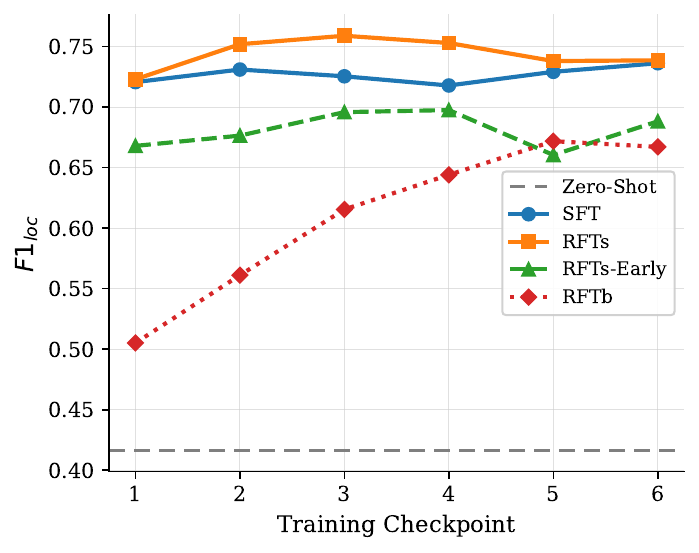}
    \caption{}
    \label{fig:ood_loc}
  \end{subfigure}
  \hfill
  \begin{subfigure}[b]{0.32\textwidth}
    \centering
    \includegraphics[width=\textwidth]{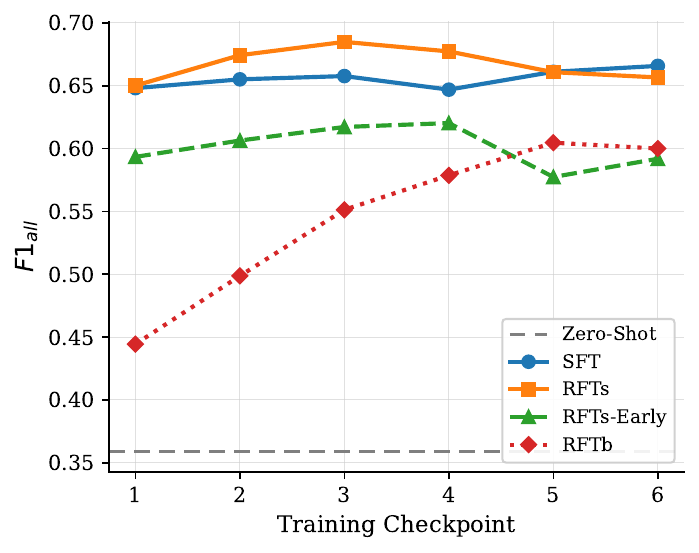}
    \caption{}
    \label{fig:ood_all}
  \end{subfigure}
  
\caption{\textbf{OOD Dynamics.} (a) \textbf{RFTb} (Red) retains the highest reading ability among trained variants, minimizing the Grounding Divergence. (b) \textbf{RFTs} (Orange) breaks the SFT Stagnation (Blue), achieving a distinct geometric surge where the supervised baseline fails to improve. \textbf{(Best viewed zoomed in.)}}
\label{fig:ood_dynamics}
\end{figure*}

To analyze how reinforcement learning interacts with supervised initialization, we study training trajectories under normalized steps on ID (Figure~\ref{fig:id_dynamics}) and DOGR-Bench OOD sets (Figure~\ref{fig:ood_dynamics}). In addition to \textbf{SFT} and \textbf{RFTs}, we introduce \textbf{RFTs-Early}, where RL is applied after only 300 SFT steps.

\paragraph{Early transition improves efficiency.}
On ID data, \textbf{RFTs-Early} achieves performance comparable to \textbf{SFT} despite using significantly less supervised data. In particular, it surpasses \textbf{SFT} in localization while matching overall grounding performance ($F1_{\text{all}}$). This indicates that transitioning to RL early is not merely a cost-saving strategy but can lead to more efficient learning of geometric alignment. Concretely, RFTs-Early uses only 300 SFT steps ($\approx$19k sample-exposures) compared to the full 1,113 SFT steps ($\approx$71k sample-exposures), then applies the same 6k RL steps as RFTs a $\sim$65\% reduction in total training data.

\paragraph{Grounding Divergence in dynamics.}
On OOD semantic extraction (Figure~\ref{fig:ood_em}), all trained variants fail to match the zero-shot base model, indicating that improvements obtained through joint optimization do not fully transfer to semantic extraction under distribution shift. Among trained models, \textbf{RFTb} retains the strongest semantic performance, suggesting that reduced specialization preserves broader generalization. In contrast, localization (Figure~\ref{fig:ood_loc}) exhibits a different trend. While \textbf{SFT} shows stagnation, \textbf{RFTs} produces a transient performance surge, reaching higher peak localization before converging. This indicates that reinforcement learning discovers transferable geometric patterns that benefit localization under distribution shift, even when semantic extraction does not improve to the same extent.
\subsection{Effect of Reasoning under RL}
\label{sec:reasoning_ablation}

\begin{table}[htbp]
\centering
\caption{\textbf{Reasoning ablation under cold-start RL.} Under identical optimization conditions, \textbf{RFTb} consistently outperforms \textbf{Reasoning-RFTb} on both ID and OOD datasets.}
\label{tab:reasoning_ablation}
\begin{tabular}{l c ccc}
\toprule
\textbf{Model} & \textbf{Dataset} & $\mathbf{F1}_{\textbf{EM}}$ & $\mathbf{F1}_{\textbf{loc}}$ & $\mathbf{F1}_{\textbf{all}}$ \\
\midrule
RFTb & ID & 0.601 & 0.496 & 0.411 \\
Reasoning-RFTb  & ID & 0.550 & 0.395 & 0.303 \\
\midrule
RFTb & OOD & 0.732 & 0.667 & 0.600 \\
Reasoning-RFTb  & OOD & 0.577 & 0.445 & 0.382 \\
\bottomrule
\end{tabular}
\end{table}

\begin{figure}[t]
\centering

\begin{subfigure}[t]{0.49\linewidth}
  \centering
  \includegraphics[width=\linewidth]{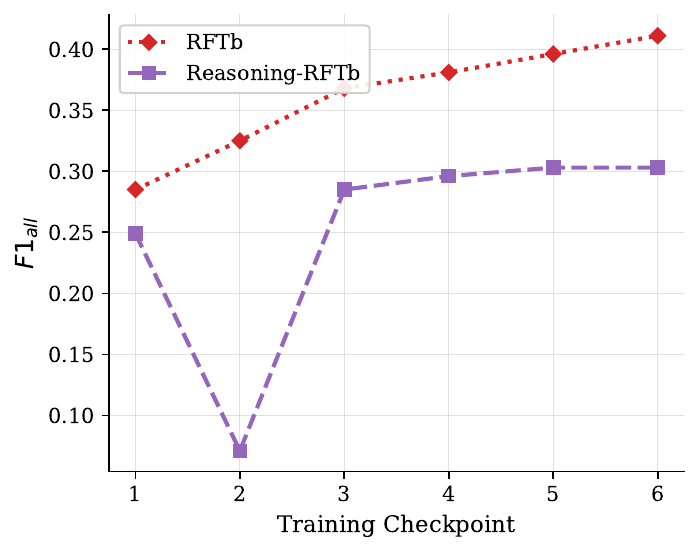}
  \caption{}
  \label{fig:reasoning_f1_iid}
\end{subfigure}
\hfill
\begin{subfigure}[t]{0.49\linewidth}
  \centering
  \includegraphics[width=\linewidth]{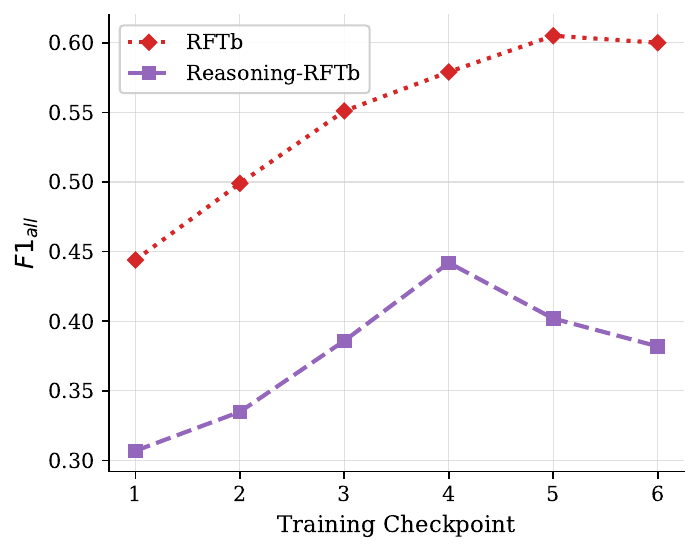}
  \caption{}
  \label{fig:reasoning_f1_ood}
\end{subfigure}

\caption{\textbf{Checkpoint-wise comparison of cold-start RL variants.} 
(\textbf{a}) ID $F1_{\text{all}}$ across checkpoints for RFTb and Reasoning-RFTb. 
(\textbf{b}) OOD $F1_{\text{all}}$ across checkpoints for the same variants. 
RFTb training shows smoother and stronger convergence, while reasoning-enabled training exhibits higher variance and inferior final performance.}
\label{fig:reasoning_f1_dynamics}

\end{figure}

\begin{figure}[htbp]
\centering
\includegraphics[width=\linewidth]{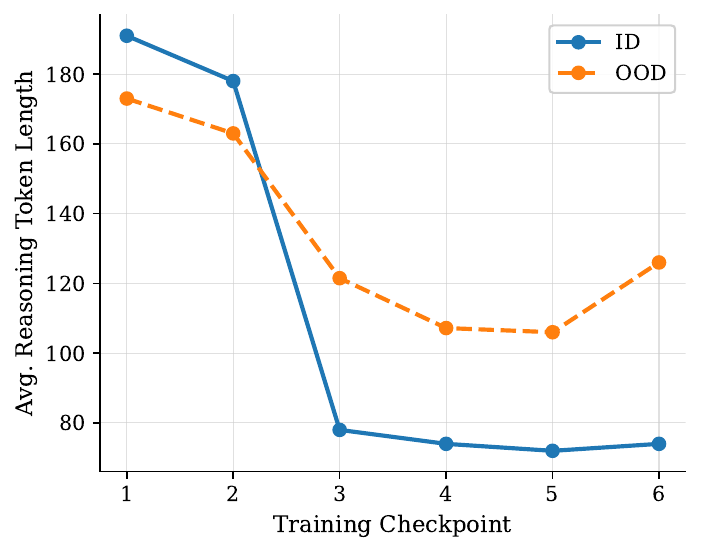}
\caption{\textbf{Reasoning token-length dynamics under cold-start RL.} 
Both ID (solid) and OOD (dashed) settings show a strong reduction in reasoning length during training. 
While ID exhibits stable compression, OOD shows a slight increase at later stages, indicating higher variability under distribution shift.}
\label{fig:reasoning_token_dynamics}
\end{figure}
We evaluate whether explicit reasoning improves document grounding under reinforcement learning by comparing \textbf{RFTb} and \textbf{Reasoning-RFTb} on ID and DOGR-Bench OOD sets in a controlled cold-start setting. This setup isolates the effect of reasoning, as reasoning-annotated supervised data are not available for constructing an SFT$\rightarrow$RFT pipeline.

\paragraph{Final performance.}
As shown in Table~\ref{tab:reasoning_ablation}, \textbf{RFTb} consistently outperforms \textbf{Reasoning-RFTb} in both ID and OOD settings. The performance gap is substantial in the joint grounding ($F1_{\text{all}}$) and is reflected in both semantic extraction and localization. This indicates that explicit reasoning does not improve the final task performance under RL.

\paragraph{Optimization behavior.}
Figure~\ref{fig:reasoning_f1_dynamics} shows that \textbf{Reasoning-RFTb} exhibits higher variance and less stable convergence. In ID data, performance drops at intermediate checkpoints before partial recovery, while \textbf{RFTb} improves steadily. In OOD data, reasoning-enabled training peaks early and then degrades, whereas \textbf{RFTb} continues to improve before stabilizing around the peak. These trends suggest that reasoning introduces additional instability during optimization.

\paragraph{Reasoning trace compression.}
We further analyze whether reasoning is retained in the learned policy by tracking the length of the output token. Figure~\ref{fig:reasoning_token_dynamics} shows a consistent reduction in reasoning length during training, particularly in the ID setting. At convergence, average reasoning token length drops from $\sim$191 to $\sim$72 tokens on ID data a \textbf{62\% reduction in per-query inference cost} directly translating to faster and cheaper deployment. This indicates that the model progressively compresses or eliminates intermediate reasoning, converging toward direct output generation.

Under identical cold-start RL conditions, explicit reasoning neither improves performance nor stabilizes training. Instead, the model converges toward a perception-driven policy, achieving higher accuracy and more consistent optimization behavior.

In practice, these findings motivate a deployment-aware choice: \textbf{RFTs} maximize joint grounding precision, \textbf{RFTb} preserves broader OOD generalization by avoiding over-specialization, and \textbf{RFTs-Early} offers the best data efficiency with comparable performance.
\section{Conclusion and Limitations}
\label{sec:conclusion}

This work studies the role of reinforcement learning in document visual grounding and introduces \textbf{Perception-RFT}, a framework for directly optimizing structured outputs without relying on intermediate reasoning. Across multiple training regimes, three key observations emerge: (1) supervised initialization provides a stable foundation for RL optimization, (2) reinforcement learning improves geometric alignment beyond the limits of likelihood-based training, and (3) across two OOD benchmarks (DOGR-Bench and MMDocBench, 4,828 samples total), these gains are not uniform, with localization improving consistently while semantic robustness exhibits task-dependent variation. A controlled reasoning ablation further shows that explicit reasoning does not improve performance or generalization under RL and is not retained in the final policy. Instead, models converge toward direct perception-based output generation, achieving higher accuracy and more stable training dynamics a finding we establish at the 4B parameter scale, with scaling behavior left as future work.

\noindent\textbf{Limitations and Future Work.}
This study focuses on a 4B parameter model, and it remains unclear how these findings scale to larger models. Additionally, the reasoning analysis is limited to cold-start RL: constructing a reasoning-enabled SFT$\rightarrow$RL pipeline would require reasoning-annotated supervised grounding data, which is not available for document visual grounding a gap we identify as a key challenge for the community. Extending this comparison to hybrid training regimes once such annotations become available is an important direction for future work. Finally, addressing the trade-off between localization and semantic robustness under distribution shift remains an open challenge.


\bibliography{example_paper}
\bibliographystyle{icml2026}

\newpage
\appendix
\onecolumn
\section{Appendix}

\subsection{Prompt Engineering}
\label{app:prompts}
We document the two system prompts used across our experimental configurations: the Direct Perception prompt (Perception-RFT), the reasoning-enabled prompt (Reasoning-RFTb)

\noindent\textbf{1. Perception-RFT System Prompt:}
\begin{quote}
\texttt{You are a high-precision visual grounding and document extraction engine. Your task is to analyze the given image and the user's question, identify the exact text in the image that answers the question, and return a tightly enclosing bounding box for that text.}
\vspace{0.5em}

\texttt{\textbf{IMPORTANT RULES:}}
\begin{itemize}[noitemsep,topsep=0pt,leftmargin=*]
  \item \texttt{Do NOT include explanations, reasoning, or intermediate steps.}
  \item \texttt{Do NOT reveal chain-of-thought.}
  \item \texttt{Output ONLY the final result.}
  \item \texttt{The output MUST be a valid JSON object.}
  \item \texttt{Do NOT include any text before or after the JSON.}
  \item \texttt{Do NOT include tags, markdown, or code fences.}
\end{itemize}
\vspace{0.5em}

\texttt{\textbf{REQUIRED OUTPUT FORMAT:}}\\
\texttt{\{"answer": "exact answer text from the image", "bbox\_2d": [x1, y1, x2, y2]\}}\\
\texttt{The bounding box must tightly enclose only the answer text.}
\end{quote}

\noindent\textbf{2. Reasoning-RFTb System Prompt:}
\begin{quote}
\texttt{You are a high-precision visual grounding and document extraction engine. Your task is to analyze the given image and the user's question, locate the visual region that answers the question, extract the exact text, and return a tightly enclosing bounding box.}
\vspace{0.5em}

\texttt{\textbf{IMPORTANT RULES:}}
\begin{itemize}[noitemsep,topsep=0pt,leftmargin=*]
  \item \texttt{You MUST first formulate your extraction plan internally. Describe your visual reasoning and localization process to identify the correct text and determine its coordinates.}
  \item \texttt{Enclose your entire reasoning process within \textless think\textgreater{} and \textless/think\textgreater{} tags.}
  \item \texttt{After thinking, provide your final result enclosed within \textless answer\textgreater{} and \textless/answer\textgreater{} tags.}
  \item \texttt{The final result inside the \textless answer\textgreater{} tags MUST be a single, valid JSON object.}
  \item \texttt{Do NOT use markdown formatting inside the \textless answer\textgreater{} tags.}
\end{itemize}
\vspace{0.5em}

\texttt{\textbf{REQUIRED OUTPUT FORMAT:}}\\
\texttt{\textless think\textgreater}\\
\texttt{Your reasoning and spatial localization process goes here...}\\
\texttt{\textless/think\textgreater}\\
\texttt{\textless answer\textgreater}\\
\texttt{\{"answer": "exact answer text from the image", "bbox\_2d": [x1, y1, x2, y2]\}}\\
\texttt{\textless/answer\textgreater}
\end{quote}

\begin{figure}[h!]
  \centering
  \includegraphics[width=0.9\textwidth]{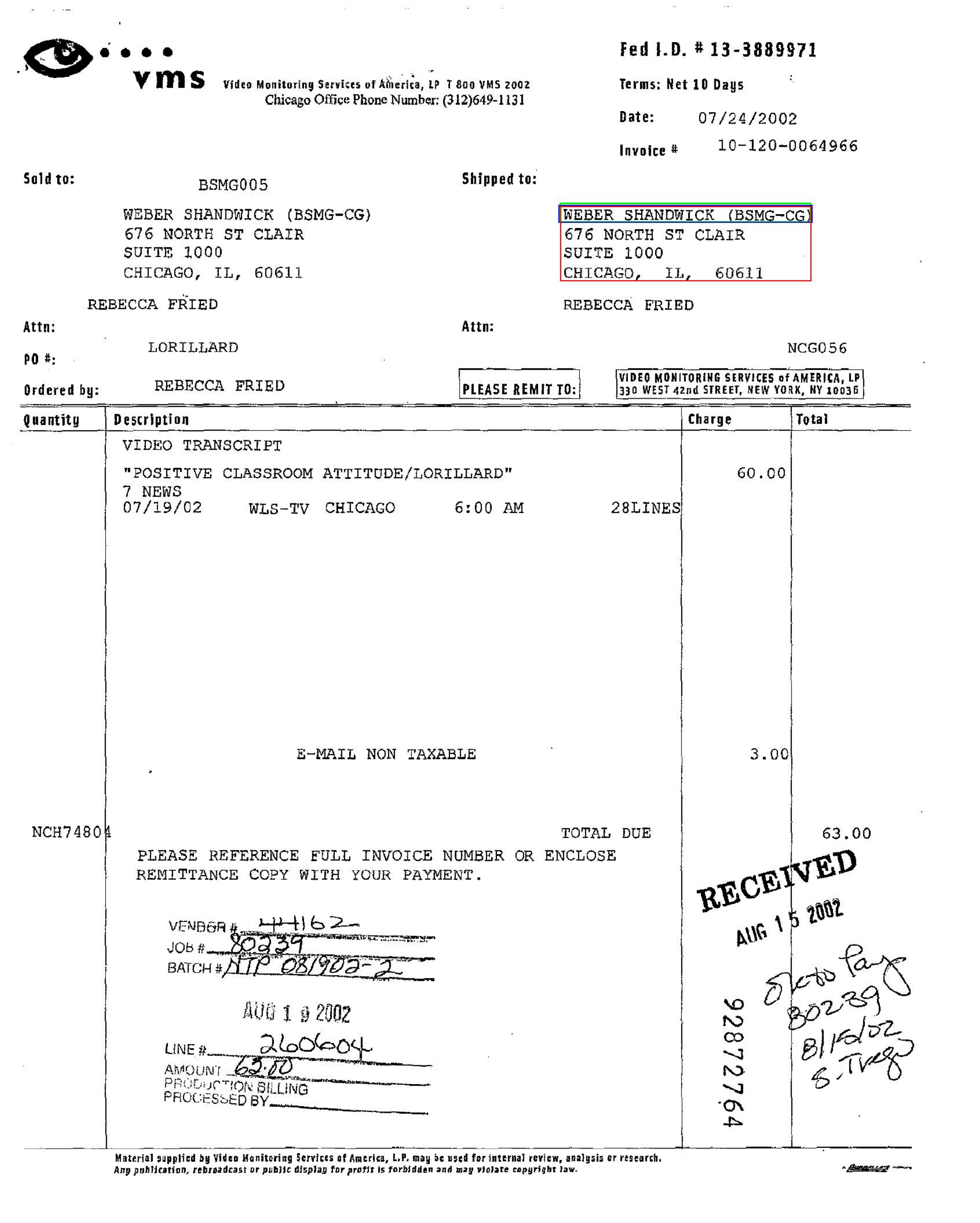}
  \caption{\textbf{Qualitative Analysis: SFT vs.\ Perception-RFT (ID).} 
  On dense financial documents, SFT often drifts into nearby words. RFT corrects this by snapping to the pixel edge. \\
  \textbf{Question:} ``According to the text, what value is given for the ship-to name?'' \\
  \textbf{\textcolor{green}{Ground Truth:}} WEBER SHANDWICK (BSMG-CG) \\
  \textbf{\textcolor{red}{SFT:}} WEBER SHANDWICK (BSMG-CG)\\676 NORTH ST CLAIR\\SUITE 1000\\CHICAGO, IL, 60611 (IoU 0.21 - Loose Box) \\
  \textbf{\textcolor{blue}{Perception-RFT:}} WEBER SHANDWICK (BSMG-CG) (IoU 0.88 - Pixel-Perfect)}
  \label{fig:qual_ID}
\end{figure}


\end{document}